\documentclass[10pt, a4paper]{article}
\usepackage{xspace}
\newcommand*{\project}{iDEM\@\xspace}
\usepackage{makecell}

\usepackage[final]{lrec2026} 
\usepackage{soul}
\usepackage{booktabs}

\title{A Multilingual Human Annotated Corpus of Original and Easy-to-Read Texts to Support Access to Democratic Participatory Processes}

\name{Stefan Bott\textsuperscript{1}, Verena Riegler\textsuperscript{2}, Horacio Saggion\textsuperscript{1}, \\
        \textbf{\large Almudena Rascón Alcaina\textsuperscript{3}, Nouran Khallaf\textsuperscript{4}}} 

\address{\textsuperscript{1}Universitat Pompeu Fabra (Barclona), \textsuperscript{2}Capito AI, \textsuperscript{3}Plena Inclusión Madrid, \textsuperscript{4}University of Leeds \\
         \{stefan.bott, horacio.saggion\}@upf.edu, verena.riegler@capito.ai,\\ 
         almudenarascon@plenamadrid.org, n.khallaf@leeds.ac.uk\\}

\abstract{Being able to understand information is a key factor for a self-determined life and society. It is also very important for participating in democratic processes. The study of automatic text simplification is often limited by the availability of high quality material for the training and evaluation on automatic simplifiers. This is true for English, but more so for less resourced languages like Spanish, Catalan and Italian. In order to fill this gap, we present a corpus of original texts for these 3 languages, with high quality simplification produced by human experts in text simplification. It was developed within the iDEM project to 
assess the impact of Easy-to-Read (E2R) language for democratic participation.
The original texts were compiled from domains related to this topic. The corpus includes different text types, selected based on relevance, copyright availability, and ethical standards. All texts were simplified to E2R 
level. The corpus is particularity valuable because it includes the first annotated corpus of its kind for the Catalan language. It also represents a noteworthy contribution for Spanish and Italian, offering high-quality, human-annotated language resources that are rarely available in these domains. The corpus will be made freely accessible to the public.
\newline }

\begin{document}

\maketitleabstract

\section{Introduction}
\label{sec:intro}

Accessibility and linguistic inclusion are increasingly recognized as critical concerns in various settings within society. Ensuring accessibility and linguistic inclusion is a growing priority in both public communication but also increasingly required by law. The European Accessibility Act (EAA) 2025 \footnote{Directive (EU) 2019/882 of the European Parliament and of the Council of 17 April 2019 on the accessibility requirements for products and services  \href{https://eur-lex.europa.eu/eli/dir/2019/882/oj}{https://eur-lex.europa.eu/eli/dir/2019/882/oj}} mandates that all public sector digital content be accessible, requiring compliance with Web Content Accessibility Guidelines (WCAG) 2.1 Level AA standards\footnote{Web Content Accessibility Guidelines 2.1, W3C World Wide Web Consortium Recommendation \href{https://www.w3.org/TR/WCAG21/}{https://www.w3.org/TR/WCAG21/}}. This includes providing content that is clear, concise, and easily navigable for users with disabilities, with a focus on multilingual adaptation and cultural relevance. 

Linguistic barriers significantly hinder participation and inclusion, according to the UNESCO Institute for Statistics, a total of 750 million adults are either illiterate or have low literacy levels\footnote{UNESCO Institute for Statistics. (2017). Factsheet No. 45: Literacy rates continue to rise from one generation of the next [Fact sheet].}. Amongst them are persons with disabilities, who account for 15\% of the world population, and have been identified as having a higher risk of illiteracy\footnote{UNESCO Institute for Statistics. (2018). Education and disability: analysis of data from 49 countries.}. According to the Special Olympics\footnote{\href{https://www.specialolympics.org/about/press-releases/special-olympics-calls-on-governments-to-commit-3-percent-of-education-funding-for-students-with-intellectual-disabilities}{https://www.specialolympics.org/about/press-releases/special-olympics-calls-on-governments-to-commit-3-percent-of-education-funding-for-students-with-intellectual-disabilities}}, persons with intellectual disabilities represent 1 to 3\% of the world population (75 to 225 million people). Moreover, only in the fourth quarter of 2018, migrants accounted for 144,166 arrivals of non-EU citizens to Europe: this group will be at a disadvantage when they have to interact with democratic institutions through accessing and providing information in a language they still do not master.


Deliberation and democratic participation is a very important area where linguistic accessibility barriers have an especially detrimental effect. People with reading difficulties are not simply less informed, they become largely invisible to democratic processes because the lack of information access leads to a lack of their formulation of their needs. Many political decisions will affect them, regardless of whether they are heard before those decisions are made. The matters of deliberation are often complex, but it is at least feasible to remove some of the linguistic difficulties that prevent persons from taking part in the deliberation. 

A traditional and proven way of facilitating inclusion, which is fortunately becoming more frequent, is the provision of simplified texts produced by human E2R translators. This has also been facilitated by the laws we mentioned above. But human intervention is expensive and usually limited to privileged contents. A potential solution in the area of powerful Language Models is an automatization of text simplification.

State-of-the-art text simplification systems can mitigate this problem by the use of modern generative AI models. But, even if Large Language Models encode much general language knowledge and exhibit surprising capabilities to solve linguistic tasks they have not been explicitly trained for (like the task of text simplification), LLMs still need data for fine-tuning and system evaluation.

The availability of data is the key factor for progress in the area. Existing E2R corpora have laid foundational work. 
There remains, however, a significant gap in multilingual E2R resources, particularly in politically relevant domains, which exemplify clear and simple language to improve text accessibility. 
Further on, there is a clear scarcity of high-quality data that comprises well-selected original text and simplified versions of it that is produced by experts. This scarcity is both one of high-quality data and data in certain languages. For Catalan, there is no available data, so far, that exemplifies sentence-level text simplification, while for Spanish and Italian, the data-pool is also very limited.
We address this gap here. 



We developed annotated E2R corpora in Spanish, Catalan, and Italian, focusing on political, deliberative, and participatory contexts. Texts were simplified to a level equivalent to A2, beginner-level proficiency in the Common European Framework of Reference for Languages (CEFR), following 
our own curated 
annotation schema and adapted to local standards (e.g., UNE). In this paper we present the work carried out in the iDEM project \cite{SaggionEtal2024} towards the creation of the iDEM Corpus.
Our contributions are as follows:

\begin{itemize}
    \item We developed of a cross-lingual methodology for human simplification and annotation.
    \item We created the first annotated E2R corpus for Catalan and corpora for Italian and Spanish which are substantial additions to the data pool.
    \item The data provided is of high quality. Simplifications were produced by trained human experts with experience in the field of E2R simplification.
    \item The simplifications were based on a well-defined list of simplification criteria for the editors. 
    \item The translators annotated which of the criteria they applied. The annotations are bundled with the corpus.
    \item The data will be publicly available.
\end{itemize}

In the next section (\S \ref{sec:rel_work}) we briefly review E2R approaches and language resources in the area of text simplification, then in \S \ref{sec:design}  we describe the corpus design choices and implementation, including its structure, formatting, distribution, and annotation process.  We continue in \S \ref{sec:corpus_details}, we provide details of the unique characteristics of the dataset and follow in \S \ref{sec:limitations} with its limitations. Finally, we conclude the paper in \S \ref{sec:conclusion} with a recap of the main contributions and impact of this new multilingual dataset in the area of text simplification.  

\section{Related Work}
\label{sec:rel_work}

\subsection{Easy-to-Read Language}

The field of written content accessibility has produced several methods for text adaptation over time, notably Plain Language \cite{plainLanguage,daCunha'2022,daCunha'2021} and Easy-to-Read (E2R)\cite{maas_easy_2020-1,Matamala&Garcia'21}. E2R is considered the most accessible variant of the two, particularly suited for individuals with intellectual disabilities \cite{Garcia'2013}. Features include the application of linguistic text simplification principles and  adoption of  design principles such as  short lines, large typography, supplementary imagery, and explanations for complex terminology \cite{nomura_guidelines_2010}. Recommendations and standards for E2R adaptation have been published by organizations including IFLA \cite{nomura_guidelines_2010} and Inclusion Europe  \cite{inclusion_europe_information_2009}, and available standards \cite{ISO24495-1:2023,ISO/IEC_23859:2023}. Despite the existence of these established standards, there is a notable scarcity of empirical research evaluating the effectiveness of these E2R guidelines \cite{gonzalez-sorde_empirical_2024}.

\subsection{Easy-to-Read Language and Text Simplification Resources}

Recent advances  in Natural Language Processing (NLP) have led to the development of various corpora aimed at enhancing text accessibility \cite{martin2023review}, particularly in political and civic domains. However, multilingual datasets adhering to E2R standards remain limited.


There are several English-language corpora for text simplification, like the Simple English Wikipedia \citep{Coster&Kauchak2011} and Newsela \citep{Xu-EtAl:2015:TACL}, but high-quality resources are still scarce. Some datasets are not fully parallel and only align comparable sentences, like in the case of the different Wikipedia based corpora \citep{kajiwara-komachi-2016-building,xu2016optimizing,zhang-lapata-2017-sentence}. In this case sentences are paired which roughly, but not fully, contain the same content. Other corpora, like ASSET \citep{alva-manchego_asset_2020} or the HIT-MTurk corpus \citep{Xu-EtAl:2015:TACL} have simplified versions produced by cloud-workers. These are no experts in E2R and tend to produce minimal solutions because of time pressure. Other corpora, still, are only aligned on the document level \citep{crossley2012text,zaman2020htss} or contain only simplified texts \citep{hauser2022multilingual} without the original versions.

Comparable datasets in Spanish, Catalan, and Italian are scarce. \citet{saggion2015making} created a first Spanish parallel corpus of non-simplified and simplified texts\footnote{This dataset is available on a one-to-one basis by contacting the corpus creators.}.
Saggion et al. \citeyearpar{saggion2024lexical} notably introduced datasets in Spanish and Catalan, marking significant contributions to the field. But these datasets are for lexical simplification and only contain simpler alternatives for individual words, not fully simplified sentences. Perez-
Rojas et al. \shortcite{perez2023novel} introduced the FEINA dataset, which contains 5314 complex and simplified sentence pairs using established simplification rules. 
For Italian there are two popular datasets: The Terrence/Teacher corpus \citet{brunato2015design} and the SIMPITIKI corpus \citet{tonelli2016simpitiki}. It must be stressed that for Catalan no dataset exists for full sentence simplification that contains parallel pairs of original and simplified texts.

Another noteworthy problem are restrictive licenses for some of the datasets with high quality simplifications, very frequently because of copyright restrictions of the source material. Many resources are not publicly available and others are only available on request. Especially the latter often become unavailable over time because of institutional changes and fluctuation of personnel. These were also some of our primary concerns in the creation of our resources.


\section{Corpus Design and Creation}
\label{sec:design}

The corpus, consisting of three language specific sub-corpora, was designed with a flexible structure to support a wide range of research and application needs. From the outset, they were developed with the explicit goal of being openly available to the scientific community. This design ensures compatibility with diverse use cases in accessibility research, multilingual NLP, and language simplification studies.

Text selection of the original, non-simplified, texts was closely aligned with the objectives of the iDEM project. The consortium collaboratively selected texts through internal voting, prioritizing materials that best reflected the project’s goals and testing requirements. The resulting corpora cover a variety of domains---government communication, news, political discourse, and legislative texts---providing a representative basis for research on democratic participation, deliberation, and inclusive communication.

\subsection{A Comprehensive Annotation Schema}
\label{subsec:annotation_scheme}

The annotation scheme was central to the creation of the corpus. 
This scheme consists of a set of rules for each language which has to be used by the E2R translators when they work.
So it is not only used for annotation, but is also the basis for the creation of the simplifications. The translators had to apply the rules from the scheme when they produced simplified sentences, identify the rule and annotate which rule they applied. 

Quality assurance was one of our main concerns. In this multilingual setting, a central challenge was the integration of three distinct languages while ensuring that quality standards remained comparable across them. Each language presents its own specific needs and complexities, and even within a single language or country, opinions often diverge regarding E2R guidelines and the appropriate design of simplification processes. Addressing this gap was one of our primary objectives. The outcome is a unified mapping of the E2R criteria for each language, developed to provide a consistent and transparent framework.
The core requirements and annotation criteria were informed by international standards as well as by extensive experience in commercial (manual) text simplification. Two of our project partners 
brought decades of expertise and teams of trained language professionals to the initiative. This collaboration is particularly important, as the umbrella term E2R is not a protected designation and currently lacks publicly available quality standards. 

A major difficulty in maintaining consistent quality stemmed from the cross-border, multilingual nature of the project. There were no uniform guidelines or qualifications that could be universally applied to the simplification process or to translator selection across all three languages. To address this issue, we developed a universal set of mapped criteria for each language. This mapping drew upon our own 
already existing annotation schema, which has been elaborated over years, and was further refined through expert consultation, particularly for Spanish and Catalan. These consultations incorporated established frameworks such as the Spanish Standardisation UNE\footnote{
Una Norma Española (One Spanish Norm), created by the Asociación Española de Normalización \href{https://www.une.org}{https://www.une.org}} standard for Spanish and the resources of DINCAT\footnote{\href{https://dincat.cat}{https://dincat.cat}} (the Catalan Federation of Intellectual Disability) for Catalan materials.
The corpus provides fine-grained annotations for the proposed simplifications, making it a unique resource for sentence classification and rewriting into E2R.

As we said, the different languages have different needs with respect to simplification. For example, Catalan uses weak pronouns\footnote{Weak pronouns are unstressed clitics which have to appear attached to a verb, like \textit{\textbf{m}'agrada} (\textbf{I} like it).} with high frequency, in contrast to the other two languages. So, there is a simplification rule which states to not use more than one weak pronoun together. Other recommendations apply to all of the languages. An important recommendation is to avoid unfamiliar words, which can be broken down into the use of frequent verbs, the, frequent and short nouns, commonly used adjectives, the avoidance of technical terms and so on. When no simpler synonym is available, the explanation of words and technical terms is recommended. 
As for the syntax, the recommendations for all languages include the use of generic \textit{subject-verb-object} (SVO) word order, to separate different logial parts of the sentence with periods or commas and the avoidance of infrequent punctuation characters.

Table~\ref{tab:corpus_examples} features three simplification examples. 
The Catalan fragment, about the functioning of the Barcelona's Town Hall, shows transformations which favor the use of shorter sentences to convey different information chunks (i.e., 1:4 alignment), the selection of E2R synonyms such as the use of  {\em cambiar o corregir (change or correct)} instead of  {\em rectificar (rectify)} and turns an example within parentheses into an independent sentence {\em un nom mal escrit es pot cambiar (a misspelled name can be changed)}

The Italian sentence on the UN's rights of the children has been simplified by applying orthographic, syntactic, lexical, and formatting E2R specifications as follows: the acronym {\em ONU (United Nations)} is explained as well as the technical term {\em convention}, i.e. {\em è come un accordo (is like an agreement)}. The verb {\em riconoscere (recognize)} is rephrased into a simpler construction with the expression {\em ci sono (there are)}. Finally, the enumeration of rights of the children is itemized and rephrased, e.g. {\em vita priva de discrimazioni (life free of discrimination)} into {\em vivere senza essere trattati male (live without being treated badly)}. 

The Spanish example on current concerns about hate speech in the European Union illustrates the application of transformations related to the reordering of syntactic components, the use of simple tenses for verbs, the simplification of the vocabulary,  and the starting of new sentences in new lines (formatting). The past perfect form {\em ha presentado (has presented)} is transformed into {\em ya hizo (already did)} while the expression {\em aleatoriamente (randomly)} is transformed into {\em al azar (at random)} and the long and complicated expression {\em debatir las distintas causas y factores que incitan al odio (debate the different causes and factors that incite hatred)} into the more direct one {\em debatir sobre el odio (debate about hate)}

\begin{table*}[t]
    \centering
  
    {\scriptsize
    \begin{tabular}{lp{2.5in}p{2.5in}}
        \toprule
        \textbf{Language} & \textbf{Original} &  \textbf{Simplified}\\ 
        \midrule

        Catalan & L'Ajuntament de Barcelona posa a disposició de la ciutadania mitjans que permeten rectificar les seves dades personals (per exemple, un nom mal escrit) o actualitzar-les (per exemple, l’adreça postal), sens perjudici del dret de rectificació reconegut a la normativa de protecció de dades. &  L’Ajuntament de Barcelona dona la possibilitat a la ciutadania 
de canviar o corregir les seves dades personals.\newline Per exemple, si el nom està mal escrit, es pot arreglar. \newline
També es pot actualitzar informació com per exemple l’adreça postal.  \newline
Tot això sense perdre el dret per corregir  les dades.

\\        \midrule

Italian & La Convenzione ONU sui Diritti dell’Infanzia e Adolescenza riconosce a tutti i bambini e ragazzi i diritti fondamentali, come il diritto alla protezione, all’educazione e a una vita priva di discriminazioni.  &
C'è la Convenzione ONU sui Diritti dell'Infanzia e Adolescenza.  \newline
Una convenzione è come un accordo. \newline  ONU sta per Organizzazione delle Nazioni Unite. \newline  È un'unione di quasi tutti i paesi del mondo. \newline
Nella convenzione, ci sono i diritti di tutti i bambini e tutti i ragazzi. \newline Per esempio, hanno questi diritti: \newline
* essere al sicuro \newline
* studiare \newline
* vivere senza essere trattati male \\
 \midrule
Spanish & El panel europeo de ciudadanos sobre la lucha contra el odio en la sociedad ha presentado sus recomendaciones a la Comisión Europea.
Durante 3 fines de semana, 150 ciudadanos seleccionados aleatoriamente, en representación de los 27 Estados miembros de la UE, se han reunido para debatir las distintas causas y factores que incitan al odio en la UE y los diferentes ámbitos de actuación para combatirlos.  &
El panel dedicado a lucha contra el odio, \newline
ya hizó sus recomendaciones a la Comisión Europea. \newline
150 ciudadanos de la Unión Europea, \newline
elegidos al azar, \newline
estuvieron reunidos durante 3 fines de semana \newline
para debatir sobre el odio en la Unión Europea \newline
y como solucionarlo. \\

\bottomrule
    \end{tabular}
    }
    \caption{\footnotesize{Corpus Examples}}
      \label{tab:corpus_examples}
\end{table*}

\subsection{Corpus Design}
\label{subsec:design}

For all the three language specific \mbox{(sub-)}corpora we designed a common strategy for their compliation, simplification and annotation. A total of 12 public, social and private institutions were involved in the creation of the corpus. 
The annotation guidelines were collaboratively developed and specifically tailored for these corpora, respecting the needs of all three languages and the target domains. 
All texts were manually simplified and annotated  by trained experts in E2R, with cross-checks among annotators and translators to ensure consistency and quality.
Translators created a simplification for each sentence in a spreadsheet; alignments were thus produced alongside the simplifications
The corpus is not parallel across languages, but have parallel original and simplified text which are aligned on the sentence level. 
They are domain-specific and reflect authentic linguistic variation.
Text types include (among others) informative, political, and news articles, as well as policy, legislative, and social service documents
The selection criteria emphasized relevance, copyright clearance, and ethical compliance.
Data are provided in JSON format, with one file per language containing integrated metadata.
Files are modular and interoperable, allowing for combination while maintaining shared design principles and annotation criteria.

\subsection{Corpus Structure}
\label{subec:corpus_structure}

Each entry includes the following main components, whose structure is reflected in the JSON. format:

The \textit{Metadata} section 
contains general information about the document, including a unique \texttt{document\_id}, language code (\texttt{language}), and detailed metadata.
The metadata fields cover source information (e.g., website URL), linguistic and textual characteristics (e.g., number of characters, corpus category, text type, topic), simplification details (level, methodology, copyright status), and information about the responsible translator and institution.

Each \texttt{text\_it} covers a text chunk of about 500 characters without spaces. Containing sets of continuous sentences are also provided for convenience. Each full document in the dataset may include multiple of these chunks, each identified by a \texttt{text\_id}. This enables comparisons of original and simplified content at the sentence, text, and document levels.

\texttt{Original and Simplified Sentences} are stored as objects with unique IDs and associated \texttt{text\_id}. Each simplified sentence can include annotations specifying complex components and applied simplification criteria. Since a single text is roughly 500 characters (without spaces), longer documents are split into several texts, allowing hierarchical analysis.

\texttt{Sentence Alignments} 
link original sentences with their simplified counterparts with \texttt{basic\_alignments}  and document the criteria used for simplification; 
\texttt{alignments} map original sentence IDs to simplified sentence IDs, indicating their position in the sequence and total number of resulting simplified sentences.
The presence of \texttt{text\_id} allows tracking simplification across sentences, within a single text, and across the entire document.

This structure facilitates multi-level analyses of text simplification strategies, supporting linguistic, cognitive, and usability research.

\begin{table*}[t]
    \centering
    {\scriptsize
    \begin{tabular}{lp{4in}}
        \toprule
        \textbf{Language} & \textbf{Topic} \\ 
        \midrule

Catalan & 
Sustainability \& Participation;
Social Policy \& Disability Accessibility; 
Data Protection;
Social Services \& Taxation;
Digital Governance;
Public Policy Discussion;
Social Justice;
Health \& Governance;
Justice \& Governance;
Civic Participation \\

Italian & 

Children's Rights \& Advocacy;
Civic Engagement \& Social Responsibility;
EU Economic Policy;
EU Elections; 
Organizational Law \& Statute;
Civic Participation; 
Legislative Reform \& Policy;
Referendum \& Electoral Systems;
Social Policy :
Economic \& Political Analysis;
Social Policy \& Advocacy;
Government/NGO Policy Communication;
Legislative Reform ("Taglia leggi" law);
EU Institutions;
Environmental Policy \& Local Governance \\

Spanish &  Social Policy;
Health \& Governance;
EU Economic Policy;
Justice \& Governance;
Political Activism;
Justice \& Legal Affairs;
Communism \& Ideology;
EU Legislation;
Civic Participation;
EU Elections;
EU Institutions \\

\bottomrule
    \end{tabular}
    }
    \caption{\footnotesize{Corpus' Topics}}
    \label{tab:topics}
\end{table*}

\begin{table}[h]
    \centering
    {\scriptsize
    \begin{tabular}{lrr}
        \toprule
        \textbf{Category} & \textbf{Words} & \textbf{Percentage} \\ 
        \midrule
        Informative texts & 4892 & 41.94\% \\
        Political \& Ideological articles & 1146 & 9.82\% \\
        News articles & 3817 & 32.72\% \\
        Policy \& Legislative Documents & 1810 & 15.52\% \\
        \midrule
        \textbf{Total} & \textbf{11665} & \textbf{100.00\%} \\ 
        \bottomrule
    \end{tabular}
    }
    \caption{\footnotesize{Percentage per category: Spanish}}
    \label{tab:percent_es}
\end{table}

\begin{table}[h]
    \centering    
    {\scriptsize
    \begin{tabular}{lrr}
        \toprule
        \textbf{Category} & \textbf{Words} & \textbf{Percentage} \\ 
       \midrule
        Informative texts & 6090 & 52.64\% \\
        Social Justice \& Public Policy Analysis & 1652 & 14.28\% \\
        Policy \& Administrative Documents & 3828 & 33.09\% \\
        \midrule
        \textbf{Total} & \textbf{11570} & \textbf{100.00\%} \\ 
        \bottomrule
    \end{tabular}
    }    
    \caption{Percentage per category: Catalan}
    \label{tab:percent_it}
\end{table}


\begin{table}[h]
    \centering 
    {\scriptsize
    \begin{tabular}{lrr}
        \toprule
        \textbf{Category} & \textbf{Words} & \textbf{Percentage} \\
        \midrule
        Informative Texts & 4425 & 42.56\% \\
        Institutional \& Legal Documents & 1247 & 11.99\% \\
        Political \& Legal Analysis & 2680 & 28.24\% \\
        Advocacy \& Social Responsibility  & 1789 & 17.21\% \\
        \midrule
        \textbf{Total} & \textbf{10398} & \textbf{100.00\%} \\
        \bottomrule
    \end{tabular}
    }    
    \caption{Percentage per category: Italian}
      \label{tab:percent_it}
\end{table}

\begin{table}[h]
    \centering
    {\scriptsize
    \begin{tabular}{lrrrrrr} 
     \toprule
    \textbf{Language} & \multicolumn{2}{c}{\textbf{Sentences}} & \multicolumn{2}{c}{\textbf{Words}} & \multicolumn{2}{c}{\textbf{Words per}} \\
     & \multicolumn{2}{c}{\textbf{Segments}} & \multicolumn{2}{c}{} & \multicolumn{2}{c}{\textbf{Sentence}} \\
     & \multicolumn{2}{c}{} & \multicolumn{2}{c}{} & \multicolumn{2}{c}{\textbf{Segment}} \\
    \midrule
    & \textbf{Orig} & \textbf{Simp} & \textbf{Orig} & \textbf{Simp} & \textbf{Orig} & \textbf{Simp} \\ 
    \midrule 
    Spanish &  354 & 1290 & 11665  & 10883 & 32.95 & 8.44\\
    Catalan &  380 & 405 & 11570 & 14279 & 41.32 & 35.26 \\
    Italian &  325 & 718 & 10398 & 12230 & 31.99 & 17.03\\
    \bottomrule
    \end{tabular}
    }    
    \caption{Distribution of number of sentence segments and words, and the word per sentence segment ratio for each language}
    \label{tab:corpus_sentences}
\end{table}

\begin{table*}[h]
    \centering 
    {\scriptsize
    \begin{tabular}{lrr}
        \toprule
        \textbf{Simplification Criteria} & \textbf{Count} & \textbf{Percentage}\\
        \midrule
Use of paragraph breaks or line breaks to separate ideas into different sentences & 241 & 22.8\% \\
Use of frequent/common verbs & 116 & 11.0\% \\
Use of short, simple, and commonly used nouns & 111 & 10.5\% \\
Reordering of sentences into logical structure: subject + verb + predicate & 84  & 7.9\%  \\
Elimination or substitution of abstract terms, foreign words, vague content, and/or technical jargon & 79  & 7.5\%  \\
Use of indicative mood and simple tenses (present, simple future, simple past, and imperfect) & 48  & 4.5\%  \\
Definition of technical or complex terms necessary for the document & 45  & 4.3\%  \\
Addition of subject to avoid ellipsis & 41  & 3.9\% \\

        \bottomrule
    \end{tabular}
    }    
    \caption{Most frequently used annotation tags: Spanish}
      \label{tab:annotations_es}
\end{table*}

\begin{table*}[h]
    \centering 
    {\scriptsize
    \begin{tabular}{lrr}
        \toprule
        \textbf{Simplification Criteria} & \textbf{Count} & \textbf{Percentage}\\
        \midrule

Choose words that are easily understandable and familiar to the target group. & 261 & 55.30\% \\
Only use common and frequent adverbs. Avoid adverbs ending in -ment. & 48  & 10.17\% \\
When listing elements related to a single idea in one sentence, use commas to separate them.  & 34  & 7.20\%  \\
Write quantities and figures using numbers instead of words. & 22  & 4.66\%  \\
Create a glossary or include glosses in the document if there are terms that are difficult to understand. & 22  & 4.66\%  \\
The most common punctuation marks may be used: period (.), comma (,), question mark (?),  & 19  & 4.03\%  \\
\ \  exclamation mark (!), and the at symbol (@). &   &   \\
Prioritize simple syntax. Use the neutral order of the elements of the sentence. Prefer short sentences & 15  & 3.18\%  \\
Avoid using percentages when addressing people with significant comprehension difficulties  & 10  & 2.12\%  \\

        \bottomrule
    \end{tabular}
    }    
    \caption{Most frequently used annotation tags: Catalan}
      \label{tab:annotations_ca}
\end{table*}

\begin{table*}[h]
    \centering 
    {\scriptsize
    \begin{tabular}{lrr}
        \toprule
        \textbf{Simplification Criteria} & \textbf{Count} & \textbf{Percentage}\\
        \midrule
Use of commonly used verbs & 292 & 13.34\%\\
Use of periods or commas to separate sentences with multiple ideas & 271 & 12.38\%\\
Elimination or modification of abstract terms, foreignisms, vague content, and/or technical jargon  & 264 & 12.06\%\\
Use of short, simple, and commonly used nouns & 255 & 11.65\%\\
Use of commonly used adjectives & 163 & 7.45\%\\
Use of commonly used adverbs & 143 & 6.53\%\\
Avoid use of semicolons (;), ellipses (…), parentheses, quotation marks, and other unusual punctuation & 77  & 3.52\%\\
Definition of technical or complex terms necessary for the text & 72 & 3.29\%\\
        \bottomrule
    \end{tabular}
    }    
    \caption{Most frequently used annotation tags: Italian}
      \label{tab:annotations_it}
\end{table*}

\begin{table}[h]
    \centering
    {\scriptsize
    \begin{tabular}{lrr} 
     \toprule
\textbf{Alignment} & \textbf{Count} & \textbf{Percentage} \\ \hline
\midrule 

1:1  & 587  &  55.43 \\ 
1:2  & 139  &  13.13 \\ 
1:3  & 116  &  10.95 \\ 
1:4  & 88  &  8.31 \\ 
1:5  & 49  &  4.63 \\
1:$>$5  & 80  &  7.55 \\ 
\bottomrule
\end{tabular}

    }    
    \caption{Alignments over the whole corpus}
    \label{tab:corpus_alignments}
\end{table}

\subsection{Accessing the Corpora}
A public GitHub repository 
will be established to provide access to the corpora files. 
The repository is scheduled for release in summer 2026. The publication was held back because the data will be used for the evaluation of the MER-TRANS IBERLEFT shared
task\footnote{\href{https://lastus-taln-upf.github.io/mertrans-iberlef-2026/}{https://lastus-taln-upf.github.io/mertrans-iberlef-2026/}}. 
The simplification guidelines for Spanish and Italian are IP of Capito\footnote{\href{https://www.capito.eu/}{https://www.capito.eu/}}. The guidelines for Catalan and the annotation tagsets  will be released with the corpus. 
We encourage active engagement with and use of the data, as well as feedback that may contribute to further refinement of the core dataset. As previously noted, this is the first resource of its kind in Catalan. Given that the current version was developed with a limited number of data points, expanding the dataset would be highly beneficial to enable broader applicability and more comprehensive research outcomes.

\subsection{Simplification and Annotation Process}

Translator selection in a multilingual, cross-national project such as ours entails several potential risks that can affect the reliability and comparability of the resulting corpus. These risks include differences in translators’ educational backgrounds, professional experience, and familiarity with linguistic simplification within political discourse. Additional challenges arise from coordinating multiple institutions across countries, as variations in local practices, terminologies, and interpretations of guidelines can introduce inconsistencies in the application of annotation and simplification protocols.
To mitigate these risks, the project relied exclusively on reputable and experienced institutions—either consortium members or long-standing collaborators with a demonstrable record of excellence in E2R work. All selected translators shared a clear understanding of the project’s annotation framework and quality assurance protocols. This approach established a common foundation that strengthened internal consistency and enhanced cross-linguistic comparability within the corpora. Three translators worked on Spanish, two on Catalan and another two on Italian. 

During the translation process, the translators annotated which of the simplification criteria they applied. So, the annotation was directly integrated with the translation workflow. The aligments resulted from the sentence-by-sentence nature of the translation. The high cost per translation and annotation, and our limited budget, did not allow for translating and annotating the corpus by more than one translator. However, we sampled random units from the texts and cross-checked them between different translators.


\section{Details of the Corpus per Language}
\label{sec:corpus_details}

Table \ref{tab:topics} shows the topics of the texts for each of the languages. All topics center around central issues in democracy, topics which are usually subject to public deliberation, and deliberation and participation themselves. 

Table \ref{tab:corpus_examples} shows some examples of original and simplified texts. E2R guidelines recommend that each main idea should be placed on a single line. This can be seen in the Italian and the Spanish examples. The Catalan editors did not follow this recommendation, but these additional segmentations are being added and will be included in the release of the corpus. Note that this recommendation adds additional newlines and somewhat blurs the notion of \textit{sentence} in the grammatical sense. We will therefore talk here about \textit{sentence segments}, a technical term which describes these sentence-like units that cut across what we traditionally understand as sentences.

The Tables \ref{tab:percent_es} to \ref{tab:percent_it} describe the distribution of text categories for each language. The text \textit{categories} are different from the \textit{topics} shown in Table \ref{tab:topics}. While the \textit{topic} describes the content, the \textit{category} describes the format of the publication. Both attributes are part of the metadata section of each corpus text.

Table \ref{tab:corpus_sentences} shows the count of words and sentence segments for each language. It also shows the average number of words per sentence segment. In general, there are more sentence segments in the simplified versions and fewer words per sentence segment. This table also reflects a difference in how much the translators have followed the E2R recommendation to place each main idea in a separate line in different languages. The Spanish translators followed it very strictly and placed only 8.44 words on average on each line, while the Catalan translators did not follow it and 'only' reduced the average length of sentences by roughly 6 words from an average of 41.32 to 35.26. For Italian the numbers are between these two extremes. Table \ref{tab:corpus_alignments} show to how many new segments the sentences have been transformed. A little bit more than half of the sentences were only translated to a single new sentence and 13.13\% were transformed to two new sentences. It is interesting that as much as 7.55\% of the sentences were transformed to 5 or more segments in the simplified version. This proliferation often corresponds to list items, but also to very complex original sentences which contained a high number of subordinate and coordinate clauses. The workflow did not foresee many-to-one or many-to-many relations, as well as deletions (1:0 alinments) between original and simplified sentences and they were excluded by design. 

The Tables \ref{tab:annotations_es} to \ref{tab:annotations_it} show which simplification rules have been used for each of the languages with the highest frequency. The full list of annotation categories can be found in Appendix A. As we said above in section \ref{subsec:annotation_scheme}, the guidelines for each language had to respect different linguistic needs, and also had to harmonize between different practical E2R simplification traditions, many of the frequent simplification operations are the same for the three languages. A first big group of frequently used simplification recommendations across languages is the avoidance of abstract terms and technical jargon and the use of familiar words. This is complemented by the recommendation to use frequent, short and simple nouns, verbs, adverbs and adjectives. A second clearly distinguishable cluster are recommendations refering to syntax, either recommending to separate different ideas in the sentence by line breaks, periods and commas or reorder the words to the canonical SVO order. The Spanish translators also frequently changed verb tenses to indicative or simple tenses and the addition of a subject to avoid ellipsis. 

\section{Challenges and Limitations}
\label{sec:limitations}

As discussed in section \ref{subsec:annotation_scheme}, a major challenge for projects of this nature lies in the absence of official guidelines and standardized definitions, particularly across linguistic and national boundaries. While high-quality standards are typically associated with human-annotated corpora, such annotation inevitably introduces a degree of human bias and error. Language is inherently dynamic and often subject to diverse and subjective interpretations.
In this project, these factors were mitigated through the adoption of a unified annotation framework established prior to data collection and the controlled selection of senior simplification experts. 

An important consideration is that these language professionals have been engaged in text simplification for many years and are therefore unaccustomed to explicitly annotating their own processes. Much of their expertise is tacit, and articulating tacit knowledge presents a significant challenge. Consequently, this unfamiliar task of annotation may have introduced certain inconsistencies. We are still firmly convinced that the use of these professional translators is by far the best way to produce high quality data for text 
simplification. 

The selection of texts was constrained by copyright restrictions as well as certain exclusions due to ethical reasons. Nevertheless, within the project consortium, constructive dialogues and discussions emerged regarding the development of annotation criteria, ultimately enhancing the overall quality of the corpora. The opportunity to gain an external perspective on another organization’s human simplification process proved both rare and highly valuable.

\section{Conclusion and Impact}
\label{sec:conclusion}

In this paper we described our efford to produce a high-quality mulilingual corpus for Easy-to-Read text simplification. We presented the challenges we encountered, especially through the mulilingual setting we work in, and the measures to took to ensure the best quality we were able to produce with the resources we had. Most significantly, we developed the annotation scheme and produced the manual simplifications with the help of trained and very experienced professionals in the field. This sets our corpus apart from already existing resources which were often created by lay persons or by aligning comparable, but not strictly parallel data.\footnote{Our corpus is parallel on the document level, but not parallel across languages as we describe in section \ref{subsec:design}.}

As a result, we can present the one-of-its-kind multilingual, human-annotated E2R corpus for political and participatory discourse in Spanish, Catalan, and Italian. This resource is distinctive in both its composition and scale. It is the first resource for sentence simplification available in Catalan and a substantial addition to the scarce data-pool in Spanish and Italian.

The development of such a project demands substantial specialized resources and the collaboration of multiple domain experts. Leveraging the infrastructure and support provided by our  project enabled us to create these corpora and to make them accessible for the broader research community. E2R should serve as a gateway — a means of enabling wider participation and access to opportunities, whether in political engagement or other domains.

We carefully selected texts from the domains of political participation and deliberative democracy. This domain requires simplification resources to a special degree because quite large parts of the population are in danger of being excluded (and are, in fact, frequently excluded) from democratic participation because of the linguistic barriers that are posed by the texts that are usually encountered in this domain. Even if we cannot mitigate the complexity of the matters of debate, we can do a lot to make the debates themselves linguistically more accessible.

The introduction of these corpora constitutes a significant advancement in promoting E2R research and fostering greater awareness of its societal relevance. Sound empirical research relies on robust data foundations, yet publicly available resources of this nature — particularly in Catalan — remain exceedingly scarce. To our knowledge, few, if any, comparable datasets combine high-quality human annotations with the scale and accessibility offered here. Consequently, this work represents a substantial contribution to the scientific community.
We also recognize that this effort marks only an initial step. The corpora may be further expanded, refined for specific applications, or utilized as evaluation datasets in future research initiatives.
Through these efforts, our project not only contributes to the development of multilingual E2R corpora but also advances inclusive communication and multilingual natural language processing, aligning with the European Accessibility Act's requirements for accessible digital content.

\section*{Acknowledgments}
This document is part of a project that has received funding from the European Union’s Horizon Europe research and innovation program under Grant Agreement No. 101132431 (iDEM Project). The views and opinions expressed in this document are solely those of the author(s) and do not necessarily reflect the views of the European Union. Neither the European Union nor the granting authority can be held responsible for them. The University of Leeds (UOL) was funded by UK Research and Innovation (UKRI) under the UK government’s Horizon Europe funding guarantee (Grant Agreement No. 10103529).

We thank  Ministerio de Ciencia, Innovación y Universidades, Agencia Estatal de Investigaciones: project CPP2023-010780 funded by MICIU/AEI/10.13039/501100011033 and by FEDER, UE (“Habilitando Modelos de
Lenguaje Responsables e Inclusivos”), the Spanish State Research Agency under the Maria de Maeztu Units of Excellence Programme (CEX2021-001195-M) and from the Departament de Recerca i Universitats de la Generalitat de Catalunya.

\section{Bibliographical References}
\bibliographystyle{lrec2026-natbib}
\bibliography{LREC-CORPUS/idem_biblio}

\section{Language Resource References}
\label{lr:ref}

\onecolumn
\section*{Appendix A: Complete List of Annotation Tags per Language and their Frequencies}

\begin{table*}[ht!]
    \centering 
    {\scriptsize
    \begin{tabular}{lrr}
        \toprule
        \textbf{Simplification Criteria} & \textbf{Count} & \textbf{Percentage}\\
        \midrule
Use of paragraph breaks or line breaks to separate ideas into different sentences & 241 & 22.8\% \\
Use of frequent/common verbs & 116 & 11.0\% \\
Use of short, simple, and commonly used nouns & 111 & 10.5\% \\
Reordering of sentences into logical structure: subject + verb + predicate & 84  & 7.9\%  \\
Elimination or substitution of abstract terms, foreign words, vague content, and/or technical jargon & 79  & 7.5\%  \\
Use of indicative mood and simple tenses (present, simple future, simple past, and imperfect) & 48  & 4.5\%  \\
Definition of technical or complex terms necessary for the document & 45  & 4.3\%  \\
Addition of subject to avoid ellipsis & 41  & 3.9\% \\
Avoid use of semicolons (;), ellipses (...), etc., parentheses, quotation marks, and other uncommon punctuation &      38 &     3.29\% \\
Elimination of abbreviations, acronyms, and initialisms or addition of explanations &     35 &     3.03\% \\
Use of frequent/common adjectives &      33 &     2.85\% \\
Replacement of words with small numeric figures &     20 &     1.73\% \\
Elimination of nominalizations &     19 &     1.64\% \\
Elimination of adverbs ending in -ly (mente) &      18 &     1.56\% \\
Elimination of impersonal sentences &     16 &     1.38\% \\
Use of headings to anticipate content &     15 &     1.30\% \\
Use of frequent/common adverbs &      14 &     1.21\% \\
Use of list formatting &     13 &     1.12\% \\
Elimination of polysemy or clarification with context &     12 &     1.04\% \\
Change of subject to directly address the reader &     11 &     0.95\% \\
Replacement of large numbers with qualitative comparisons or other terms &     9 &      0.78\% \\
Elimination of ellipses (as a figure of speech) &     7 &      0.61\% \\
Elimination of compound tenses &     7 &      0.61\% \\
Elimination of parenthetical elements and appositives &     7 &      0.61\% \\
Use of colons (:) to introduce lists &      6 &      0.52\% \\
Transcription &      5 &      0.43\% \\
Elimination of sarcasm, metaphors, and humor &     5 &      0.43\% \\
Elimination of complex verbal periphrases &     5 &      0.43\% \\
Elimination of gerunds &     5 &      0.43\% \\
Change to capitalization for sentence beginnings and proper nouns &      4 &      0.35\% \\
Elimination of fractions, percentages, and ordinal numbers &     3 &      0.26\% \\
Use of pronouns with clear referents &      3 &      0.26\% \\
Elimination of complex connectors and use of frequent/common ones  &     3 &      0.26\% \\
Explanation &     2 &      0.17\% \\
Elimination of double negatives &     1 &      0.09\% \\
Elimination of superlative adjectives & 0 & -\\
        \bottomrule
    \end{tabular}
    }    
    \caption{All annotation tags with frequencies: Spanish}
      \label{tab:annotations_es_full}
\end{table*}

\begin{table*}[h]
    \centering 
    {\scriptsize
    \begin{tabular}{lrr}
        \toprule
        \textbf{Simplification Criteria} & \textbf{Count} & \textbf{Percentage}\\
        \midrule
Choose words that are easily understandable and familiar to the target group. & 261 & 55.30\% \\
Only use common and frequent adverbs. Avoid adverbs ending in -ment. & 48  & 10.17\% \\
When listing elements related to a single idea in one sentence, use commas to separate them.  & 34  & 7.20\%  \\
Write quantities and figures using numbers instead of words. & 22  & 4.66\%  \\
Create a glossary or include glosses in the document if there are terms that are difficult to understand. & 22  & 4.66\%  \\
The most common punctuation marks may be used: period (.), comma (,), question mark (?),  & 19  & 4.03\%  \\
\ \  exclamation mark (!), and the at symbol (@). &   &   \\
Prioritize simple syntax. Use the neutral order of the elements of the sentence. Prefer short sentences & 15  & 3.18\%  \\
Avoid using percentages when addressing people with significant comprehension difficulties  & 10  & 2.12\%  \\
Use verbs to describe actions  &     6 &      0.98\% \\
Use weak pronouns with a clear reference   &     6 &      0.98\% \\
Use abbreviations, acronyms, and initialisms only when strictly necessary &     5 &      0.81\% \\
Use gender-neutral forms whenever possible, depending on the target group, content, scope, and publication format &      5 &      0.81\% \\
Avoid using synonymous words to refer to the same concept or referent. &     5 &      0.81\% \\
Avoid the use of words and expressions with a figurative meaning with ironic or sarcastic intention. &     3 &      0.49\% \\
Avoid infrequent verbal tense. &     2 &      0.33\% \\
Prefer affirmative sentences over negative ones, except in simple prohibitions. &     2 &      0.33\% \\
Write informative titles that refer to and anticipate the content. &      1 &      0.16\% \\
Limit content to what is essential Try not to introduce more than two ideas in the same sentence. &      1 &      0.16\% \\
Use paragraphs that span several pages without splitting them, each approximately 4 to 8 lines long. &      1 &      0.16\% \\
Visually highlight the separation between paragraphs. &  &  \\
Maintain a clear and logical structure with headings and paragraphs to ensure immediate  &      1 &      0.16\% \\
comprehension by the target group. &   &    \\
Separate phone numbers into evenly sized blocks with a space between each block. &     1 &      0.16\% \\
Avoid perspective shifts as much as possible. & 0 & -\\
Provide an overview (index) at the beginning. & 0 & -\\
Do not use cross-references. & 0 & -\\
Texts with dialogues should follow a theatrical style, indicating the character's name for each line. & 0 & -\\
Redacction and disign of summaries have to follow the rules and recommendations of the general design. & 0 & -\\
Summaries have to appear at the end of the section or chapter where the reference is made. & 0 & -\\
The main text has to be easily distingushable in order to not confuse and hinder reading. & 0 & -\\
Prefer the active voice. & 0 & -\\
Avoid comparatives and superlatives. Use "more" instead & 0 & -\\
Avoid abstract, technical, specialised and complex words. If unavoidable explain add a gloss or explanation. & 0 & -\\
Use subordinate clauses with moderation, linking them with basic and frequently used conjuctions. & 0 & -\\
Avoid complex connectors.  & 0 & -\\
Use weak preposition without restrictions. & 0 & -\\
Use only the most frequently used strong prepositions. & 0 & -\\
Use the most freuquent prepositional phrases. & 0 & -\\
Use only the most frequent modifiers. & 0 & -\\
If homophones and homographs are used with different meaning the text must clarify this. & 0 & -\\
Don't write words or sentences in upper case, except for acronyms.  & 0 & -\\
Write the full date and the name of the day when additional information is necessary to understand the text.   & 0 & -\\
Avoid the use of Roman numerals.  & 0 & -\\
        \bottomrule
    \end{tabular}
    }    
    \caption{All annotation tags with frequencies: Catalan}
      \label{tab:annotations_ca_full}
\end{table*}

\begin{table*}[h]
    \centering 
    {\scriptsize
    \begin{tabular}{lrr}
        \toprule
        \textbf{Simplification Criteria} & \textbf{Count} & \textbf{Percentage}\\
        \midrule
Use of commonly used verbs & 292 & 13.34\%\\
Use of periods or commas to separate sentences with multiple ideas & 271 & 12.38\%\\
Elimination or modification of abstract terms, foreignisms, vague content, and/or technical jargon  & 264 & 12.06\%\\
Use of short, simple, and commonly used nouns & 255 & 11.65\%\\
Use of commonly used adjectives & 163 & 7.45\%\\
Use of commonly used adverbs & 143 & 6.53\%\\
Avoid use of semicolons (;), ellipses (…), parentheses, quotation marks, and other unusual punctuation & 77  & 3.52\%\\
Definition of technical or complex terms necessary for the text & 72 & 3.29\%\\
Use of bullet point formatting &     72 &     3.29\% \\
Elimination of relative subordinate clauses &     72 &     3.29\% \\
Elimination of impersonal sentences and passive voice verbs &     64 &     2.92\% \\
Use of colons (:) to introduce lists &      58 &     2.65\% \\
Elimination of parenthetical and appositive phrases &     49 &     2.24\% \\
Elimination or explanation of abbreviations and acronyms &     48 &     2.19\% \\
Elimination of nominalizations &     43 &     1.96\% \\
Elimination of conditional or subjunctive forms &     37 &     1.69\% \\
Reorganization of the sentence in logical order: subject–verb–object &     34 &     1.55\% \\
Elimination of sarcasm, metaphors, and humor &     32 &     1.46\% \\
Elimination of the gerund &     26 &     1.19\% \\
Use of the indicative mood and simple verb tenses (present indicative, present imperative,  &     23 &     1.05\% \\
simple future, present perfect) &  &  \\
Use of digits instead of spelling out numbers &     22 &     1.01\% \\
Use of pronouns with a clear reference to the person or subject &      20 &     0.91\% \\
Use of titles to anticipate content &     17 &     0.78\% \\
Elimination of rarely used conjunctions30 &     12 &     0.55\% \\
Substitution of large numbers with qualitative comparisons or other wording &     7 &      0.32\% \\
Change in phrasing to directly address the reader &     6 &      0.27\% \\
Elimination of fractions, percentages, and ordinal numbers &     3 &      0.14\% \\
Specification of the subject &     3 &      0.14\% \\
Elimination of uncommon superlative forms &      3 &      0.14\% \\
Elimination of ellipses (as a stylistic device) &     1 &      0.05\% \\        
Change to capitalization at the beginning of sentences and for proper nouns. & 0 & - \\
\bottomrule
    \end{tabular}
    }    
    \caption{All annotation tags with frequencies: Italian}
      \label{tab:annotations_it_full}
\end{table*}

\end{document}